\documentclass{Interspeech}

\interspeechcameraready

\usepackage{inconsolata}
\usepackage{booktabs}
\usepackage{soul}
\usepackage{color}

\usepackage{multirow}
\usepackage{multicol}
\usepackage{enumitem,kantlipsum}
\usepackage[normalem]{ulem}
\usepackage{color,colortbl}
\usepackage{todonotes} 
\usepackage{soul}
\usepackage{threeparttable, makecell}
\usepackage{makecell}
\usepackage{xcolor}

\usepackage{comment}
\usepackage{array}  
\usepackage{lscape}  
\usepackage{listings} 
\usepackage{multirow}
\usepackage{graphicx}
\usepackage[normalem]{ulem}
\useunder{\uline}{\ul}{}

\usepackage{array}  
\usepackage{multirow}  
\usepackage{amsmath}
\usepackage{subcaption}
\usepackage{fancybox}
\usepackage{tikz}

\usepackage{arabtex}
\usepackage{utf8}
\setcode{utf8}

\definecolor{codegray}{gray}{0.9}

\lstdefinestyle{pythonstyle}{
    backgroundcolor=\color{codegray},   
    language=Python,
    basicstyle=\ttfamily\footnotesize,
    keywordstyle=\color{blue},
    stringstyle=\color{red},    
    breaklines=true,
    frame=single,
    keepspaces=true,
    showstringspaces=false,
}

\usepackage{graphicx}
\definecolor{lightblue}{rgb}{.50,.95,1}
\definecolor{tri}{rgb}{.25,.88,.82}
\definecolor{lilac}{rgb}{0.85,0.64,0.85}

\newcommand{\snqa}{\emph{SpokenNativQA}}

\usepackage{ulem}
\usepackage{hyperref}
\usepackage{verbatim}
\usepackage{setspace}

\usepackage{colortbl}
\usepackage{booktabs}

\title{SpokenNativQA: Multilingual Everyday Spoken Queries for LLMs}

\author[affiliation={1}]{Firoj}{Alam}
\author[affiliation={2}]{Md Arid}{Hasan$^\dagger$}
\author[affiliation={1}]{Shammur Absar}{Chowdhury}

\affiliation{}{Qatar Computing Research Institute}{Qatar}
\affiliation{}{University of New Brunswick}{Canada}
\email{\{fialam,shchowdhury\}@hbku.edu.qa, arid.hasan@unb.ca}


\usepackage{comment}

\begin{document}

\maketitle
\begingroup
\renewcommand{\thefootnote}{}
\footnotetext{$\dagger$Work done during internship at QCRI.}%
\endgroup

\begin{abstract}
Large Language Models (LLMs) have demonstrated remarkable performance across various disciplines and tasks. However, benchmarking their capabilities with multilingual spoken queries remains largely unexplored. In this study, we introduce \textit{SpokenNativQA}, the \textit{first, multilingual and culturally aligned} spoken question-answering (SQA) dataset designed to evaluate LLMs in real-world conversational settings. The dataset comprises approximately 33k naturally spoken questions and answers in multiple languages, including low-resource and dialect-rich languages, providing a robust benchmark for assessing LLM performance in speech-based interactions. \textit{SpokenNativQA} addresses the limitations of text-based QA datasets by incorporating speech variability, accents, and linguistic diversity. We benchmark different ASR systems and LLMs for SQA and present our findings. We released the data\footnote{{\href{https://huggingface.co/datasets/QCRI/SpokenNativQA}{https://huggingface.co/datasets/QCRI/SpokenNativQA}}}
and experimental scripts\footnote{{\href{https://llmebench.qcri.org/}{https://llmebench.qcri.org/}}}
 for the research community.
\end{abstract}

\section{Introduction}
\label{sec:intro}

The emergent capabilities of LLMs have significantly increased their adoption among end users, enabling seamless integration into everyday tasks. These models are now widely utilized across diverse domains, from education to professional activities, where their ability to generate human-like responses, provide contextual understanding, and support complex problem-solving has enhanced user engagement and efficiency \cite{openai2023gpt4,bubeck2023sparks,touvron2023llama}. While most interactions with LLMs are text-based, their use with spoken queries is also growing, particularly through virtual assistants.
In recent years, research in Spoken Question Answering (SQA) has gained significant attention~\cite{menevcse2022framework,zhao2024librisqa}. Traditionally, SQA systems have relied on a cascaded approach that integrates Automatic Speech Recognition (ASR) with text-based models \cite{huang2024audiogpt}. However, this approach suffers from several limitations, including error propagation from ASR outputs and poor handling of speech-specific phenomena such as disfluencies, hesitations, and prosody. With advancements in LLMs, the scientific community has begun exploring more robust, end-to-end multimodal frameworks to address these challenges~\cite{zhang-etal-2023-speechgpt,chu2023qwen,hu2024wavllm,cheng2023ghostt5}.

\begin{figure}[t]
  \centering
\includegraphics[width=1.0\columnwidth]{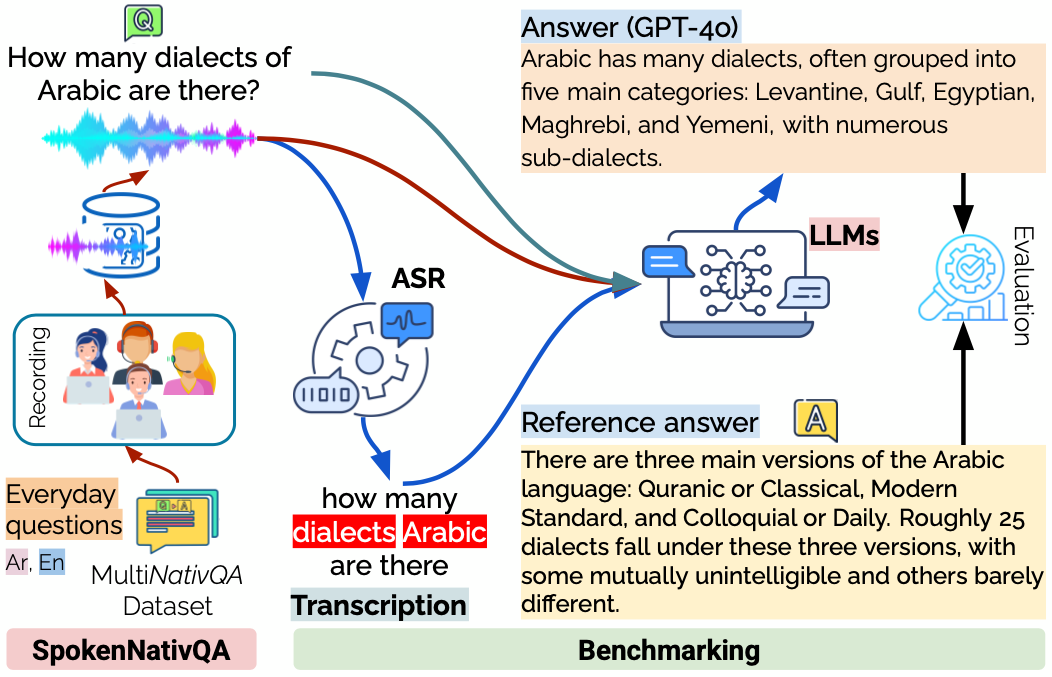}
  \caption{Complete overview of the \snqa{} dataset development pipeline and benchmarking experiments.}
  \vspace{-0.3cm}
  \label{fig:spokennativqa}
  \vspace{-0.4cm}
\end{figure}

To evaluate the effectiveness of both LLM-based and traditional SQA systems, several benchmarking  efforts have been made \cite{lee2018spoken,zhao2024librisqa,labrak2024zero,wu2023heysquad,shankar2024coraal}. For instance, Spoken SQuAD \cite{lee2018spoken} laid the foundation for understanding the impact of ASR errors on SQA performance. Subsequent datasets, such as LibriSQA \cite{zhao2024librisqa}, Spoken-CoQA \cite{you2020towards}, and HeySQuAD \cite{wu2023heysquad}, extended evaluations to more realistic, noisy, and conversational spoken environments. These datasets have facilitated benchmarking LLMs for SQA. However, current efforts in benchmarking LLMs and Multimodal LLMs (MLLMs) for SQA remain relatively limited~\cite{wang2024audiobench} compared to text-based benchmarks~\cite{liang2022holistic,bang2023multitask}. Moreover, the majority of \textbf{existing SQA datasets are primarily limited to English and are synthetic}, (see Table \ref{tab:dataset_comparison}) highlighting the need for \textbf{native, multilingual resources} that focus on everyday queries to improve spoken language understanding across diverse linguistic contexts.

To address this gap, we propose \snqa{}, a dataset focused on \textit{native}, \textit{local}, and \textit{cultural} specific everyday queries. Figure \ref{fig:spokennativqa} presents an example QA pair, illustrating a common everyday question that can be asked by both native and non-native speakers. While the question pertains to region- or location-specific information, it remains relevant to any speaker.
In this study, we leverage this dataset to evaluate the performance of current LLMs on everyday queries using a cascaded pipeline. Additionally, we compare their performance with multimodal LLMs (e.g., GPT-4 Audio). The \textit{\textbf{major contributions}} of our study are as follows:

\begin{itemize}[noitemsep,topsep=0pt,labelsep=.5em]
    \item We propose \snqa{}, featuring $\sim$30 hours of everyday questions, comprising 33,081 samples from 12 and 11 speakers in Arabic and English, respectively. 
    \item We provide a comparative evaluation of different ASR systems using our dataset.
    \item We benchmark LLMs and MLLM on everyday SQA, offering a performance comparison. 
\end{itemize}

\noindent
\textbf{\textit{Our findings}} are summarized as follows: 
\textit{(i)} ASR performance is affected for English, possibly because all speakers are L2 speakers, whereas ASR performance is comparatively better for Arabic L1 speakers. \textit{(ii)} The performance of SQA reflects the performance of ASR systems. Google location-based ASR performs relatively better, which corresponds to its matching performance with the No-ASR setup (Section \ref{sec:finding}). A similar phenomenon is observed for English with Whisper. Our findings reinforce the need for cascade-less systems, which are crucial for everyday queries, especially those that may have significant impacts, such as health-related queries.

\section{Dataset}
\label{sec:dataset}

\noindent
We adopted the MultiNativQA \cite{hasan2024nativqa} dataset, which includes QA pairs where queries come from real users and answers were manually edited. The dataset consists of questions covering 18 different topics that are culture- and region-specific, as well as everyday questions. The topics in the dataset include \textit{Animals, Business, Clothing, Education, Events, Food \& Drinks, General, Geography, Immigration, Language, Literature, Names \& Persons, Plants, Religion, Sports \& Games, Tradition, Travel}, and \textit{Weather}. 
For this study, we utilize the test set of Arabic and English (Qatar) from the MultiNativQA dataset.
The Arabic and English test sets contain 988 and 2,322 QA pairs, respectively. As reported in \cite{hasan2024nativqa}, the dataset has undergone several rounds of manual annotation. During the QA annotation phase, following the annotation guidelines, two annotators were first asked to verify whether the question was related to the location, then categorize the answer, and finally edit the answer if needed. 

\subsection{\snqa{}} To develop \snqa{}, we recruited speakers who are native in Arabic and fluent in English. We hosted an in-house recording platform to facilitate the recording process, where individual questions were displayed. Speakers were asked to record audio by reading the question in the respective language and then submit it. We did not impose any restrictions on how to read the question or on the recording environment and requirement of any specific recording devices, as the goal was to reflect a real-world setup. 
Each question was recorded by ten speakers. However, a total of 12 and 11 speakers participated in the question recording for Arabic and English, respectively. The audio was recorded in a mono-channel setup at a 48 kHz sampling rate.
The complete recording of the \snqa{} resulted in $\sim30$ hours of recording associated with $\sim33k$ samples, with an average duration of $\sim3$ seconds. To compensate the speakers we hired a third party company who managed the payment process and paid standard hourly rate. Each speaker signed a Non-Disclosure Agreement (NDA) for their participation. 

\begin{figure}[h]
  \centering
\includegraphics[width=0.75\columnwidth]{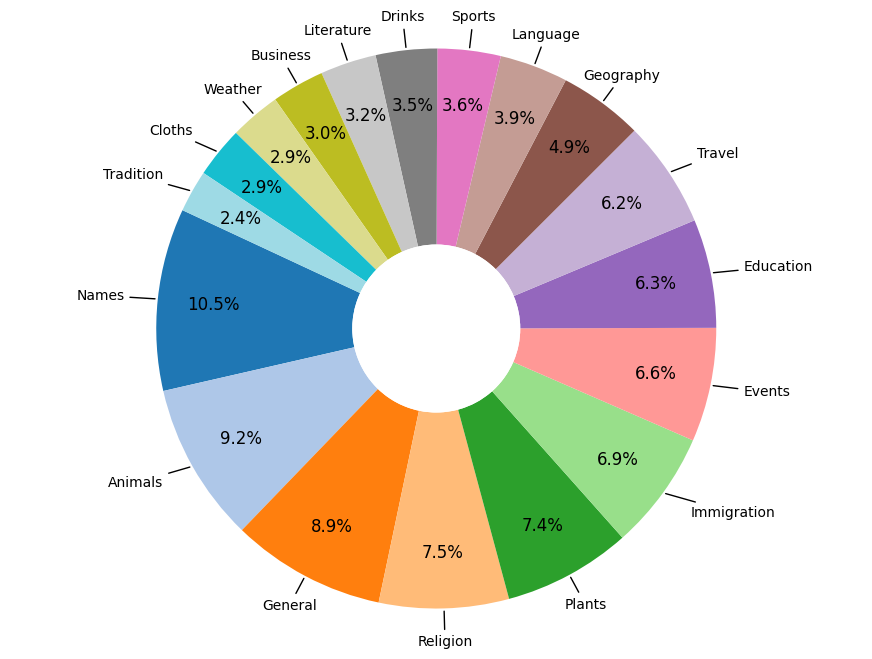}
  \caption{Topic wise distribution for Arabic.}
  \vspace{-0.3cm}
  \label{fig:ar_topic_distribution}
  \vspace{-0.1cm}
\end{figure}

\begin{figure}[h]
  \centering
\includegraphics[width=0.75\columnwidth]{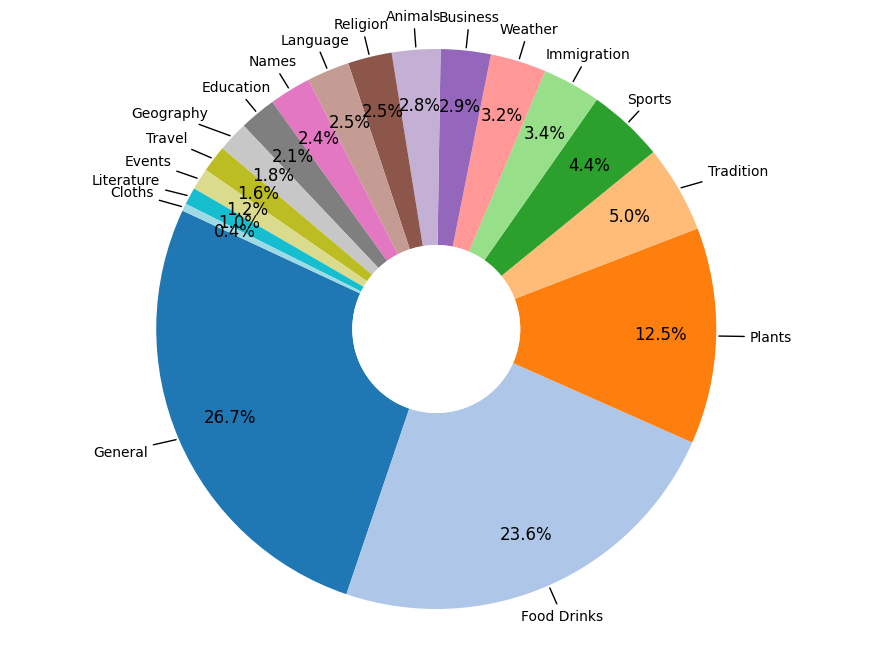}
  \caption{Topic wise distribution for English (Qatar).}
  \vspace{-0.2cm}
  \label{fig:en_topic_distribution}
  \vspace{-0.1cm}
\end{figure}

We provide topic wise distribution of Arabic and English data in Figure \ref{fig:ar_topic_distribution} and \ref{fig:en_topic_distribution}. In the Arabic data, the top 5 topics based on the number of QA pairs are names, animals, general, religion, and plants while the least data belongs to tradition, cloths, and weather topics. However, in English data, $\sim50$\% QA pairs belong to only `general' and `food drink' topics while cloths, literature, and events have 0.4\%, 1.0\%, and 1.2\% QA pairs respectively. We also provide speaker wise data distribution in Table \ref{tab:speaker_details} for both test sets where 12 speakers recorded their voices for Arabic and 11 speakers for English.

\begin{table}[h]
\centering
\setlength{\tabcolsep}{2pt} 
\scalebox{0.75}{
\begin{tabular}{lrrrrrrrrrrrrr}
\toprule
\textbf{} & \multicolumn{12}{c}{\textbf{Speakers}} &  \\ \cline{2-13}
\textbf{Lang} & \textbf{S1} & \multicolumn{1}{c}{\textbf{S2}} & \multicolumn{1}{c}{\textbf{S3}} & \multicolumn{1}{c}{\textbf{S4}} & \multicolumn{1}{c}{\textbf{S5}} & \multicolumn{1}{c}{\textbf{S6}} & \multicolumn{1}{c}{\textbf{S7}} & \multicolumn{1}{c}{\textbf{S8}} & \multicolumn{1}{c}{\textbf{S9}} & \multicolumn{1}{c}{\textbf{S10}} & \multicolumn{1}{c}{\textbf{S11}} & \multicolumn{1}{c}{\textbf{S12}} & \textbf{Total} \\ \midrule
\textbf{Arabic} & 92 & 96 & 89 & 110 & 97 & 62 & 109 & 99 & 77 & 69 & 32 & 56 & \textbf{988} \\
\textbf{English} & 232 & 250 & 225 & 204 & 87 & 240 & 221 & 241 & 152 & -- & 232 & 238 & \textbf{2,322} \\
\bottomrule
\end{tabular}
}
\caption{Speaker wise data distribution for test subset.}
\label{tab:speaker_details}
\vspace{-0.4cm}
\end{table}


\subsection{Test Subset} For this study, we selected the test subset from the full set. We randomly selected one speaker's recording for each question. Moreover, we ensured that recordings from all speakers were included in the test subset. Detailed statistics on recording duration for both the full set and the test subset are provided in Table \ref{tab:data_stat}. 

\begin{table}[h]
\centering
\setlength{\tabcolsep}{3pt} 
\scalebox{0.8}{%
\begin{tabular}{@{}lrrr@{}}
\toprule
\multicolumn{4}{c}{\textbf{Full set}} \\ \midrule
\textbf{Lang} & \multicolumn{1}{l}{\textbf{Total Dur}} & \multicolumn{1}{l}{\textbf{Avg. Dur (s)}} & \multicolumn{1}{l}{\textbf{\# of samples}} \\ \midrule
Arabic & 9h, 16m, 8.46s & 3.38 & 9877 \\
English & 21h, 40m, 28.14s & 3.36 & 23204 \\ \midrule
\multicolumn{4}{c}{\textbf{Test subset}} \\ \midrule
Arabic & 55m, 15.24s & 3.36 & 988 \\
English & 2h, 9m, 31.68s & 3.35 & 2322 \\ \bottomrule
\end{tabular}
}
\caption{Distribution of dataset.}
\vspace{-0.6cm}
\label{tab:data_stat}
\vspace{-0.2cm}
\end{table}

\section{Methodology}
As presented in Figure \ref{fig:spokennativqa}, our methodological steps consist of \snqa{} dataset development, discussed in the previous section, and benchmarking. The latter part involves employing several open and closed ASR systems to obtain transcriptions for both languages. To understand the impact of ASR error propagation, we benchmark three LLMs and a specialized LLM (GPT-4o-audio), then compare their performance with No-ASR QAs.

\subsection{ASR Systems}
As for the ASR systems we used Google (AR-QA for English), Azure (AR-QA for Arabic), Fanar~\cite{fanar2024}, and an open model -- Whisper (large-v3)~\cite{radford2023robust}. All are current state-of-the art for both languages. All systems are accessed through APIs except Whisper. In Table \ref{tab:asr_performance} we report the performance of the ASR systems with basic text normalization for Arabic. 
For Arabic, Google's Qatari ASR outperforms all other systems. Followed by Azure's QA and Fanar models. Compared to Google and Azure, Fanar and Whisper is a general model (not specific to a region), hence making them more suitable for multilingual SQA.

\begin{table}[h]
\centering
\setlength{\tabcolsep}{3pt} 
\scalebox{0.8}{%
\begin{tabular}{@{}lrrrr@{}}
\toprule
\multicolumn{1}{c}{\textbf{Lang}} & \multicolumn{1}{c}{\textbf{Google}} & \multicolumn{1}{c}{\textbf{Azure}} & \multicolumn{1}{c}{\textbf{Whisper}} & \multicolumn{1}{c}{\textbf{Fanar}} \\ \midrule
Arabic & \textbf{5.85} & 9.50 & 12.50 & 10.40 \\
English & 18.02 & 21.40 & \textbf{10.58} & 33.80 \\ \midrule
Avg & 11.94 & 15.45 & \textbf{11.54} & 22.10 \\
\bottomrule
\end{tabular}
}
\caption{WER for different ASR systems.}
\label{tab:asr_performance}
\vspace{-0.6cm}
\end{table}

\subsection{Models and Benchmarking Setups}
We experiment with both open and closed LLMs. For the closed models, we use GPT-4o\footnote{Referred to as an MLLM, supporting multiple modalities.}~\cite{achiam2023gpt} (version 2024-11-20) and Fanar~\cite{fanar2024}, while for the open models, we use ALLaM~\cite{bari2024allam}. Our setup employs zero-shot learning across all models. As presented in Figure \ref{fig:spokennativqa}, we also employ another LLM with (gpt-4o-audio-preview), which is specially designed for audio input and output. For prompting, we follow the prompt reported in a prior study (\cite{hasan2024nativqa}); however, we redesigned it after several iterations of manual checking to achieve optimal performance. 
In listing \ref{lst:prompt_for_answer}, we provide an example of a prompt. In the prompt we pass question and a length variable where question is either original question, a transcribed version or direct audio. We introduced length variable to limit the LLMs not to generate long-form answers. The value of the length variable is determined by based on the length of the answer in the dataset. To ensure reproducibility, we released the experimental scripts as a part of LLMeBench~\cite{dalvi-etal-2024-llmebench}. 

\begin{lstlisting}[style=pythonstyle,caption={Prompt for QA. The place-holders length and question refers to number of words for answer and question.},label={lst:prompt_for_answer}]
question_prompt = f"""
Please use your expertise to answer the following English question. 
Answer in English and rate your confidence level from 1 to 10.
Provide your response in the following JSON format: 
{{"answer": "your answer", "score": your confidence score}}.
Please provide JSON output only. No additional text. 
Answer should be limited to less or equal to {length} words.

Question: {question}
"""
system_prompt = """
You are an English AI assistant specialized in providing detailed 
and accurate answers across various fields. 
Your task is to deliver clear, concise, and relevant information.
"""
\end{lstlisting}

\noindent\textbf{Evaluation and Metrics.}
We evaluate the model's explanation performance on the \snqa{} test set using semantic similarity-based metric, measured by the F$_1$ score within BERTScore~\cite{zhangbertscore}. It is computed using contextual embeddings extracted from pre-trained BERT models. We leverage language-specific transformer models for embedding extraction: AraBERT (v2)~\cite{baly2020arabert} for Arabic and bert-base-uncased~\cite{devlin2019bert} for English. While BLEU and ROUGE are widely used, studies have highlighted their limitations for long-form QA tasks \cite{xu2023critical}. As a results, we rely solely on BERTScore.


\section{Results and Findings}
\label{sec:finding}

In Table \ref{tab:results_overall}, we present the performance of different models and compare No-ASR and ASR-cascaded setups. Based on the results, the No-ASR setup, which uses gold questions, consistently achieves the highest scores across all models and both languages. The average performance in this setup is 0.536 for Arabic and 0.619 for English. 
Note that our No-ASR results may not match those reported for \textit{MultiNativQA}~\cite{hasan2024nativqa} due to differences in setups and prompts used in both studies.

The results indicate a decline in performance when ASR is introduced. On average among the ASR systems, Whisper outperforms Google, Azure, and Fanar, achieving the best results for GPT-4o (0.571), Fanar (0.559), and ALLaM (0.544) -- average performance across three models. This suggests that while ASR negatively impacts overall performance, Whisper is the most effective ASR system for SQA.

Although Whisper performed relatively poorly among ASR systems for Arabic, it was the best-performing ASR model for English. Its ASR performance highlights its strong performance for English in the SQA task.

\begin{figure}[t]
  \centering
\includegraphics[width=0.65\columnwidth]{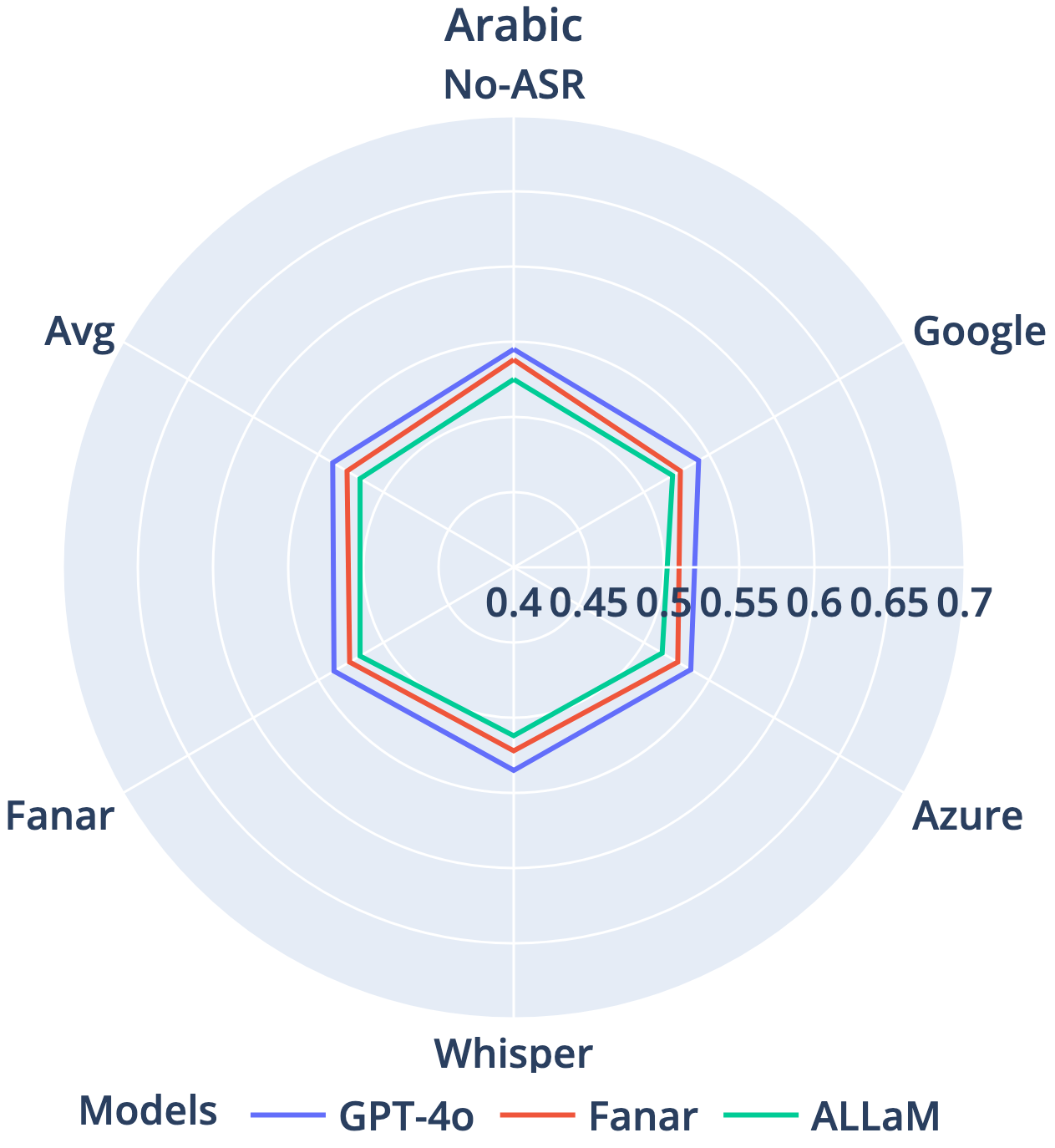}
  \caption{F1 across different setups and models for \textit{Arabic}.
  }
  \vspace{-0.3cm}
  \label{fig:results_asr_models_arabic}
\end{figure}

\noindent
\textbf{Performance for Arabic.} For Arabic, GPT-4o stands out among the models; however, its performance difference compared to Fanar is minimal ($0.011$). Overall, a lower ASR error rate (Google - 5.85) is reflected in its performance with the No-ASR setup for SQA, achieving $0.531$, which is similar to the No-ASR performance of $0.536$. 
In Figure \ref{fig:results_asr_models_arabic}, we provide a clear overview of the models' performance.

\begin{figure}[t]
  \centering
\includegraphics[width=0.65\columnwidth]{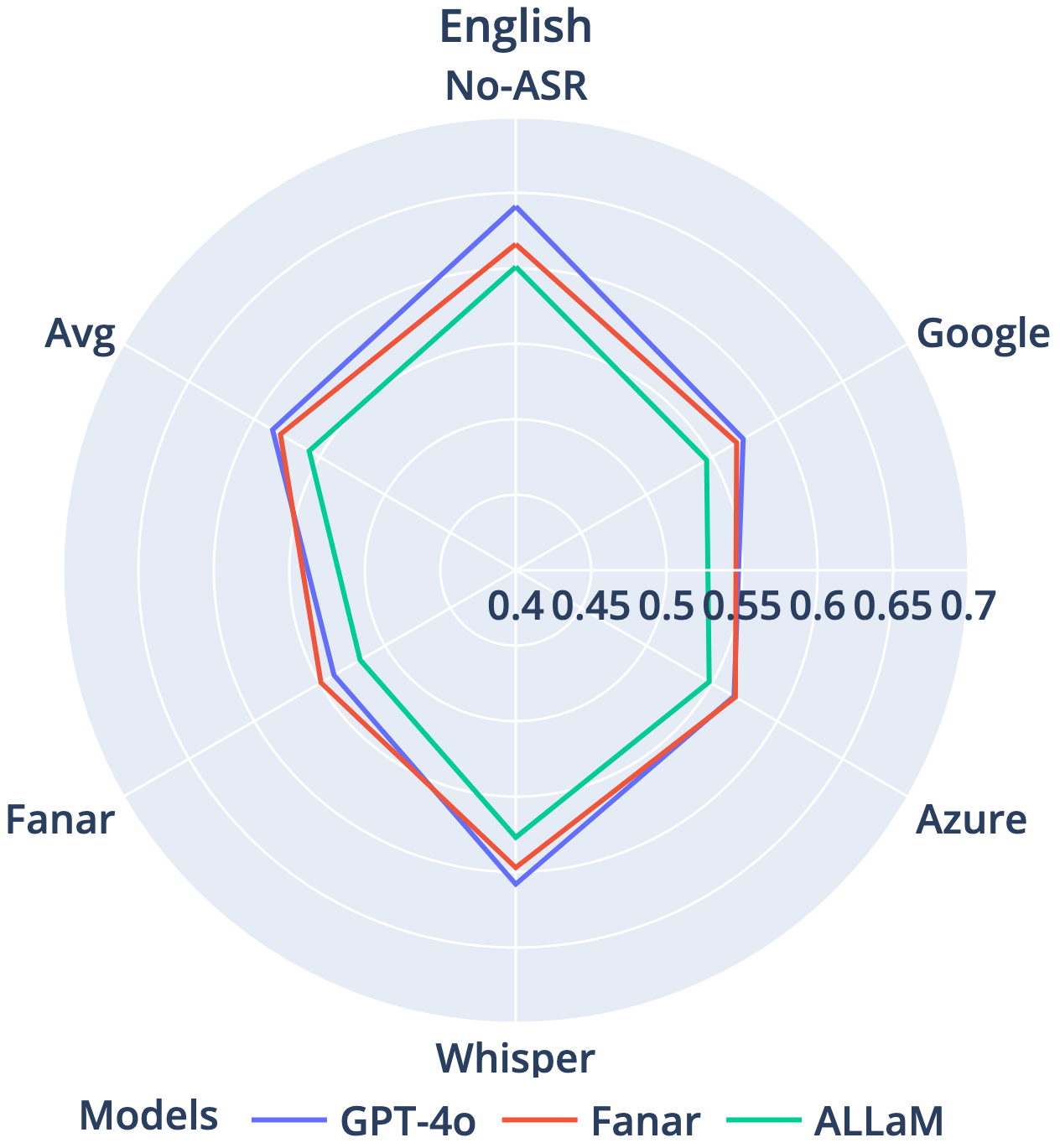}
  \caption{F1 across different setups and models for \textit{English}.}
  \vspace{-0.3cm}
  \label{fig:results_asr_models_english}
  \vspace{-0.3cm}
\end{figure}

\noindent
\textbf{Performance for English.} For English, the best performance has been achieved using Whisper ASR across all models for SQA, with almost no performance difference between GPT-4o and Fanar as reporte in Figure \ref{fig:results_asr_models_english}. 

Overall, SQA performance is significantly better for English than for Arabic, even with Arabic-centric models such as Fanar and ALLaM.

\begin{table}[]
\centering
\setlength{\tabcolsep}{2pt} 
\scalebox{0.75}{%
\begin{tabular}{@{}lcccc@{}}
\toprule
\multicolumn{1}{c}{\textbf{Setups}} & \textbf{GPT-4o} & \textbf{Fanar} & \textbf{ALLaM} & \textbf{Avg.} \\ \midrule
\multicolumn{5}{c}{\cellcolor[HTML]{EAD1DC}\textbf{Arabic}} \\ \midrule
\rowcolor{gray!20} No-ASR & 0.545 & 0.538 & 0.525 & \textbf{0.536} \\
Google & \uline{0.542} & 0.\uline{528} & \uline{0.522} & \uline{\textbf{0.531}} \\
Azure & 0.536 & 0.526 & 0.514 & 0.525 \\
Whisper & 0.535 & 0.522 & 0.512 & 0.523 \\
Fanar & 0.538 & 0.526 & 0.518 & 0.527 \\ \midrule
\textbf{Avg} & \textbf{0.539} & \textbf{0.528} & \textbf{0.518} &  \\ \midrule
\multicolumn{5}{c}{\cellcolor[HTML]{9FC5E8}\textbf{English}} \\ \midrule
\rowcolor{gray!20}  No-ASR & 0.641 & 0.616 & 0.601 & \textbf{0.619} \\
Google & 0.574 & 0.569 & 0.546 & 0.563 \\
Azure & 0.567 & 0.568 & 0.548 & 0.561 \\
Whisper & \uline{0.608} & \uline{0.597} & \uline{0.577} & \uline{\textbf{0.594}} \\
Fanar & 0.539 & 0.549 & 0.519 & 0.536 \\ \midrule
\textbf{Avg} & \textbf{0.586} & \textbf{0.580} & \textbf{0.558} &   
\\ \bottomrule
\end{tabular}
}
\caption{Performance (F1) comparison of different models with and without ASR, as well as various ASR setups. F1 refers to the score obtained from BERTScore. 
The number with \textbf{bold} and \uline{underline} refers to best across models.}
\label{tab:results_overall}
\vspace{-0.3cm}
\end{table}


\noindent
\textbf{Results on GPT-Audio.} It is unclear whether this model operates as a cascaded system. We evaluated its performance using both question and audio inputs. The F1 scores are 0.55 for Arabic and 0.62 for English. It outperforms all other models in both languages, surpassing the No-ASR setup for Arabic and matching its performance for English. The results indicate the importance of developing such a system with open models.

\noindent
\textbf{Error Analysis.} We conducted an error analysis and examined samples to understand how ASR errors impacted SQA performance. For instance, in one example, the recording of the question \textit{``What is Doha known for food?"} was transcribed by Azure ASR as \<\footnotesize وتزوها نون فورفود. >. The issue is that Azure ASR incorrectly transcribed an English utterance into Arabic, resulting in an incorrect sentence. Similarly, Fanar also transcribed it into Arabic. Only Google ASR produced a correct transcription. The incorrect transcription caused GPT-4o to respond with \textit{``The input text is not in English, so I cannot provide an answer."}, while Fanar generated a hallucinated response.

\section{Related Work} 
\label{sec:related_work}
Benchmarks play a crucial role in evaluating AI systems. For SQA, it is particularly important to assess how current LLMs and MLLMs perform under noisy, real-world conditions. A fundamental challenge in this evaluation is the availability of benchmarking datasets. Early efforts, such as Spoken SQuAD \cite{lee2018spoken}, introduced ASR-generated spoken question-answer pairs derived from the text-based SQuAD dataset, highlighting the cascading impact of ASR errors on QA accuracy. Following that, the authors in \cite{you2020towards} proposed Spoken-CoQA -- a multi-turn conversational QA benchmark, addressing challenges in maintaining context dependency over multiple interactions.
Recent datasets have increased the linguistic and contextual diversity of SQA evaluation. LibriSQA \cite{zhao2024librisqa} introduces an open-ended and multiple-choice spoken QA format, leveraging clean speech from LibriSpeech to benchmark large language models (LLMs) in multimodal reasoning. Meanwhile, HeySQuAD \cite{wu2023heysquad} extended the SQuAD QA dataset with human- and machine-generated questions, with answers sourced from SQuAD QA. 

Beyond extractive QA, CORAAL QA \cite{shankar2024coraal} is one of the first datasets to explore long-context spoken QA, utilizing spontaneous speech from sociolinguistic corpora. ASK-QA \cite{chen2024data} introduces ambiguous, diverse-accent, and prosody-aware QA to enhance robustness testing across different speaker styles. The study in \cite{lin2022dual} proposes the NMSQA benchmark, which focuses on realistic speech scenarios with specific attention to speech-specific challenges, benchmarking, and robustness under noisy conditions.
Additionally, recent efforts address low-resource languages, such as TurQuAse \cite{menevcse2022framework}, which presents a Turkish spoken QA benchmark, highlighting ASR errors and transfer learning challenges in non-English QA. 

In Table \ref{tab:dataset_comparison}, we provide a brief summary of the relevant datasets and highlight how ours differs. We explicitly focus on everyday queries, with questions recorded by Arabic L1 and English L2 speakers in realistic settings. More importantly, in \textit{SpokenNativQA}, the questions originate from real users, are recorded by humans, and the answers are human-curated. Neither the questions nor the answers are synthetic, which sets our dataset apart from prior studies. We are the \textit{first} to offer such a multilingual dataset.


\begin{table}[tbh!]
\centering
\setlength{\tabcolsep}{2pt} 
\scalebox{0.7}{%
\begin{tabular}{@{}lllll@{}}
\toprule
\multicolumn{1}{c}{\textbf{Dataset}} & \multicolumn{1}{c}{\textbf{Type}} & \multicolumn{1}{c}{\textbf{Lang}} & \multicolumn{1}{c}{\textbf{Size}} & \multicolumn{1}{c}{\textbf{SS}} \\ \midrule
Spoken SQuAD \cite{lee2018spoken} & Extractive QA & En & 100k+ & TTS \\
Spoken-CoQA \cite{you2020towards} & Conversational QA & En & 120k+ & TTS \\
NMSQA \cite{lin2022dual} & Extractive QA & En & 10.5k+ & TTS \\
TurQuAse \cite{menevcse2022framework} & Extractive QA & Tr & 30k+ & TTS \\
LibriSQA \cite{zhao2024librisqa} & Open-ended QA & En & 214k & Human \\
HeySQuAD \cite{wu2023heysquad} & Extractive QA & En & 173k & Human + TTS \\
CORAAL QA \cite{shankar2024coraal} & Long-context QA & En & 21k+ & Human \\
ASK-QA \cite{chen2024data} & Multi-turn QA & En & 7.8k+ & TTS \\
\rowcolor[HTML]{FFFC9E} 
\begin{tabular}[c]{@{}l@{}}\textbf{SpokenNativQA}\\ (Ours)\end{tabular} & \begin{tabular}[c]{@{}l@{}}Open-ended \\ every-day QA\end{tabular} & Ar, En & 33k+ & Human \\ \\ \bottomrule
\end{tabular}
}
\caption{Comparison between different SQA datasets. SS: Speech Source. Size: Number of QA Pairs.}
\label{tab:dataset_comparison}
\vspace{-0.5cm}
\end{table}

\section{Conclusion and Future Work}
In this study, we introduce a multilingual SQA dataset for everyday queries. The questions are recorded by L1 speakers for Arabic and L2 speakers for English languages. This is due to the nature of the task itself, which targets culture and region-specific QA.
We collected $\sim33k$ samples for both Arabic and English, where questions are varied over 18 different topics. We investigated the performance of different ASR systems and provided a comparison. We benchmark different LLMs for SQA both in a cascaded pipeline and with audio-specific LLMs. The results suggest the importance of end-to-end model, where error propagation can be reduced. Our future study, includes extending the dataset for other regions and locations and moving towards the development of a cascade-less LLM supporting everyday queries.

\bibliographystyle{IEEEtran}
\bibliography{bibliography/bibliography}


\end{document}